# A Game-Theoretic Analysis of Updating Sets of Probabilities

**Peter D. Grünwald**
CWI, P.O. Box 94079
1090 GB Amsterdam
pdg@cwi.nl

**Joseph Y. Halpern**
Cornell University
Ithaca, NY 14853
halpern@cs.cornell.edu

## Abstract

We consider how an agent should update her uncertainty when it is represented by a set $\mathcal{P}$ of probability distributions and the agent observes that a random variable $X$ takes on value $x$, given that the agent makes decisions using the *minimax criterion*, perhaps the best-studied and most commonly-used criterion in the literature. We adopt a game-theoretic framework, where the agent plays against a bookie, who chooses some distribution from $\mathcal{P}$. We consider two reasonable games that differ in what the bookie knows when he makes his choice. Anomalies that have been observed before, like *time inconsistency*, can be understood as arising because different games are being played, against bookies with different information. We characterize the important special cases in which the optimal decision rules according to the minimax criterion amount to either conditioning or simply ignoring the information. Finally, we consider the relationship between conditioning and *calibration* when uncertainty is described by sets of probabilities.

## 1 INTRODUCTION

Suppose that an agent models her uncertainty about a domain using a *set* $\mathcal{P}$ of probability distributions. How should the agent make decisions? Perhaps the best-studied and most commonly-used approach in the literature is to use the minimax criterion [Wald 1950; Gärdenfors and Sahlin 1982; Gilboa and Schmeidler 1989]. According to the minimax criterion, action $a_1$ is preferred to action $a_2$ if the worst-case expected loss of $a_1$ (with respect to all the probability distributions in the set $\mathcal{P}$ under consideration) is better than the worst-case expected loss of $a_2$. Thus, the action chosen is the one with the best worst-case outcome.

We are often interested in making decisions, not just in a static situation, but in a more dynamic situation, where the agent may make some observations, or learn some information. This leads to an obvious question: If the agent represents her uncertainty using a set $\mathcal{P}$ of probability distributions, how should she update $\mathcal{P}$ in light of observing that random variable $X$ takes on value $x$? Perhaps the standard answer is to condition each distribution in $\mathcal{P}$ on $X = x$ (more precisely, to condition those distributions in $\mathcal{P}$ that give $X = x$ positive probability on $X = x$), and adopt the resulting set of conditional distributions $\mathcal{P} \mid X = x$ as her representation of uncertainty. As has been pointed out by several authors, this sometimes leads to a phenomenon called *dilation* [Augustin 2003; Cozman and Walley 2001; Herron, Seidenfeld, and Wasserman 1997; Seidenfeld and Wasserman 1993]: the agent may have substantial knowledge about some other random variable $Y$ before observing $X = x$, but know significantly less after conditioning. Walley [1991, p. 299] gives a simple example of dilation: suppose that a fair coin is tossed twice, where the second toss may depend in an arbitrary way on the first. (In particular, the tosses might be guaranteed to be identical, or guaranteed to be different.) If $X$ represents the outcome of the first toss and $Y$ represents the outcome of the second toss, then before observing $X$, the agent believes that the probability that $Y$ is heads is $1/2$, while after observing $X$, the agent believes that the probability that $Y$ is heads can be an arbitrary element of $[0, 1]$.

While, as this example and others provided by Walley show, such dilation can be quite reasonable, it interacts rather badly with the minimax criterion, leading to anomalous behavior that has been called *time inconsistency* [Grünwald and Halpern 2004; Seidenfeld 2004]: the minimax-optimal conditional decision rule before the value of $X$ is observed (which has the form "If $X = 0$ then do $a_1$; if $X = 1$ then do $a_2$; ...") may be different from the minimax decision rule obtained after conditioning. For example, the minimax-optimal conditional decision rule may say "If $X = 0$ then do $a_1$", but the minimax-optimal decision rule conditional on observing $X = 0$ may be $a_2$. (See Example 2.1.) If uncertainty is modeled using a single distribution, such time inconsistency cannot arise.

To understand this phenomenon better, we model the decision problem as a game between the agent and a bookie. It turns out that there is more than one possible game that can be considered, depending on what information the bookie has. We focus on two (closely related) games here. In the first game, the bookie chooses a distribution from $\mathcal{P}$ before the agent moves. We show that the Nash equilibrium of this game leads to a minimax decision rule. (Indeed, this can be viewed as a justification of using the minimax criterion). However, in this game, conditioning on the information is not always optimal.[1] In the second game, the bookie gets to choose the distribution *after* the value of $X$ is observed. Again, in this game, the Nash equilibrium leads to the use of minimax, but now conditioning *is* the right thing to do.

If $\mathcal{P}$ is a singleton, the two games coincide (since there is only one choice the bookie can make, and the agent knows what it is). Not surprisingly, conditioning is the appropriate thing to do in this case. The moral of this analysis is that, when uncertainty is characterized by a set of distributions, if the agent is making decision using the minimax criterion, then the right decision depends on the game being played. The agent must consider if she is trying to protect herself against an adversary who knows the value of $X = x$ when choosing the distribution or one that does not know the value of $X = x$.

In earlier work [Grünwald and Halpern 2004] (GH from now on), we essentially considered the first game, and showed that, in this game, conditioning was not always the right thing to do when using the minimax criterion. Indeed, we showed there are sets $\mathcal{P}$ and games for which the minimax-optimal decision rule is to simply ignore the information. Our analysis of the first game lets us go beyond GH here in two ways. First, we characterize exactly when it is minimax optimal to ignore information. Second, we provide a simple sufficient condition for when conditioning on the information is minimax optimal.

Ignoring the information can be viewed as the result of conditioning; not conditioning on the information, but conditioning on the whole space. This leads to a natural question: suppose that when we observe $x$, we condition on the event that $X \in \mathcal{C}(x)$, where $\mathcal{C}(x)$ is some set containing $x$, but not necessarily equal to $\{x\}$. Is this variant of conditioning, an approach we call $\mathcal{C}$-*conditioning*, always minimax optimal in the first game? That is, is it always optimal to condition on *something*? As we show by considering the well-known Monty Hall Problem (Example 5.3), this is not the case in general. Nevertheless, $\mathcal{C}$-conditioning has some interesting properties: it is closely related to the concept of *calibration* [Dawid 1982]. Calibration is usually defined in terms of empirical data. To explain what it means, consider an agent that is a weather forecaster on your local television station. Every night the forecaster makes a prediction about whether or not it will rain the next day in the area where you live. She does this by asserting that the probability of rain is $p$, where $p \in \{0, 0.1, \ldots, 0.9, 1\}$. How should we interpret these probabilities? The usual interpretation is that, in the long run, on those days at which the weather forecaster predict probability $p$, it will rain approximately $100p\%$ of the time [Dawid 1982]. Thus, for example, among all days for which she predicted $0.1$, the fraction of days with rain was close to $0.1$. A weather forecaster with this property is called *calibrated*.

---
[1]In some other senses of the words "conditioning" and "optimal," conditioning on the information *is* always optimal. This is discussed further in Section 6.

Up to now, calibration has been considered only when uncertainty is characterized by a single distribution. We generalize the notion of calibration to our setting, where uncertainty is characterized by a set of distributions. We then show that a rule for updating a set of probabilities is guaranteed to be calibrated if and only if it is an instance of $\mathcal{C}$-conditioning. In combination with our earlier results, this implies that if calibration is considered essential, then an update rule may sometimes result in decisions that are not minimax optimal.

Both the idea of representing uncertainty by a set $\mathcal{P}$ of distributions and that of handling decisions in a worst-case optimal manner may, of course, be criticized. While we do not claim that this is necessarily the "right" or the "best" approach, two of the most common criticisms are, to some extent, unjustified. First, since it may be hard for an agent to determine the precise boundaries of the set $\mathcal{P}$, it has been argued that "soft boundaries" are more appropriate. While this is sometimes the case, hard boundaries are natural in some cases, such as the Monty Hall problem (Example 5.3). Similarly, the use of the minimax criterion is not as pessimistic as is often thought. The minimax solution often coincides with the Bayes-optimal solution under some "maximum entropy" prior [Grünwald and Dawid 2004], which is not commonly associated with being overly pessimistic. In fact, in the Monty Hall problem, the minimax-optimal decision rule coincides with the solution usually advocated, which requires making further assumptions about $\mathcal{P}$ to reduce it to a singleton.

## 2 NOTATION AND DEFINITIONS

**Preliminaries:** For ease of exposition, we assume throughout this paper that we are interested in two random variables, $X$ and $Y$, which can take values in spaces $\mathcal{X}$ and $\mathcal{Y}$, respectively. $\mathcal{P}$ always denotes a set of distributions on $\mathcal{X} \times \mathcal{Y}$; that is, $\mathcal{P} \subseteq \Delta(\mathcal{X} \times \mathcal{Y})$, where, as usual, $\Delta(S)$ denotes the set of probability distributions on $S$. For ease of exposition, we assume that $\mathcal{P}$ is a closed set; this is a standard assumption in the literature that seems quite natural in our applications, and makes the statement of our results simpler. If $\Pr \in \Delta(\mathcal{X} \times \mathcal{Y})$, let $\Pr_{\mathcal{X}}$ and $\Pr_{\mathcal{Y}}$ denote the marginals of $\Pr$ on $\mathcal{X}$ and $\mathcal{Y}$, respectively.

Let $\mathcal{P}_\mathcal{Y} = \{\Pr_\mathcal{Y} : \Pr \in \mathcal{P}\}$. If $E \subseteq \mathcal{X} \times \mathcal{Y}$, then let $\mathcal{P} \mid E = \{\Pr \mid E : \Pr \in \mathcal{P}, \Pr(E) > 0\}$. Here $\Pr \mid E$ (denoted by some authors as $\Pr(\cdot \mid E)$) is the distribution on $\mathcal{X} \times \mathcal{Y}$ obtained by conditioning on $E$.

**Loss Functions:** As in GH, we are interested in an agent who must choose some action from a set $\mathcal{A}$, where the loss of the action depends only on the value of random variable $Y$. For ease of exposition, we assume in this paper that $\mathcal{X}$, $\mathcal{Y}$, and $\mathcal{A}$ are always finite. We assume that with each action $a \in \mathcal{A}$ and value $y \in \mathcal{Y}$ is associated some loss to the agent. (The losses can be negative, which amounts to a gain.) Let $L : \mathcal{Y} \times \mathcal{A} \to I\!R$ be the loss function.[2]

Such loss functions arise quite naturally. For example, in a medical setting, we can take $\mathcal{Y}$ to consist of the possible diseases and $\mathcal{X}$ to consist of symptoms. The set $\mathcal{A}$ consists of possible courses of treatment that a doctor can choose. The doctor's loss function depends only on the patient's disease and the course of treatment, not on the symptoms. But, in general, the doctor's choice of treatment depends on the symptoms observed.

**Decision Rules:** Suppose that the agent observes the value of a variable $X$ that takes on values in $\mathcal{X}$. After having observed $X$, she must perform an act, the quality of which is judged according to loss function $L$. The agent must choose a *decision rule* that determines what she does as a function of her observations. We allow decision rules to be randomized. Thus, a decision rule is a function $\delta : \mathcal{X} \to \Delta(\mathcal{A})$ that chooses a distribution over actions based on the agent's observations. Let $\mathcal{D}(\mathcal{X}, \mathcal{A})$ be the set of all decision rules. A special case is a deterministic decision rule, which assigns probability 1 to a particular action. If $\delta$ is deterministic, we sometimes abuse notation and write $\delta(x)$ for the action that is assigned probability 1 by the distribution $\delta(x)$. Given a decision rule $\delta$ and a loss function $L$, let $L_\delta$ be the random variable on $\mathcal{X} \times \mathcal{Y}$ such that $L_\delta(x,y) = \sum_{a \in \mathcal{A}} \delta(x)(a) L(y, a)$. Here $\delta(x)(a)$ stands for the probability of performing action $a$ according to the distribution $\delta(x)$ over actions that is adopted when $x$ is observed. Note that in the special case that $\delta$ is a deterministic decision rule, $L_\delta(x,y) = L(y, \delta(x))$.

A decision rule $\delta^0$ is *a priori minimax optimal* with respect to $\mathcal{P}$ and $\mathcal{A}$ if

$$\max_{\Pr \in \mathcal{P}} E_{\Pr}[L_{\delta^0}] = \min_{\delta \in \mathcal{D}(\mathcal{X}, \mathcal{A})} \max_{\Pr \in \mathcal{P}} E_{\Pr}[L_\delta].$$

That is, $\delta^0$ is a priori minimax optimal if $\delta^0$ gives the best worst-case expected loss with respect to all the distributions in $\Pr$. We can write max here instead of sup because of our assumption that $\mathcal{P}$ is closed. This ensures that there is some $\Pr \in \mathcal{P}$ for which $E_{\Pr}[L_{\delta^0}]$ takes on its maximum value.

---
[2]We could equally well use utilities, which can be viewed as a positive measure of gain. Losses seem to be somewhat more standard in this literature.

A decision rule $\delta^1$ is *a posteriori minimax optimal* with respect to $\mathcal{P}$ and $\mathcal{A}$ if, for all $x \in \mathcal{X}$ such that $\Pr(X = x) > 0$ for some $\Pr \in \mathcal{P}$,

$$\begin{aligned}\max_{\Pr \in \mathcal{P}|X=x} E_{\Pr}[L_{\delta^1}] = \\ \min_{\delta \in \mathcal{D}(\mathcal{X},\mathcal{A})} \max_{\Pr \in \mathcal{P}|X=x} E_{\Pr}[L_\delta].\end{aligned} \quad (1)$$

To get the a posteriori minimax-optimal decision rule we do the obvious thing: if $x$ is observed, we simply condition each probability distribution $\Pr \in \mathcal{P}$ on $X = x$, and choose the action that gives the least expected loss (in the worst case) with respect to $\mathcal{P} \mid X = x$. Since all distributions $\Pr$ mentioned in (1) satisfy $\Pr(X = x) = 1$, the minimum over $\delta \in \mathcal{D}(\mathcal{X}, \mathcal{A})$ does not depend on the values of $\delta(x')$ for $x' \neq x$; the minimum is effectively over randomized actions rather than decision rules.

As the following example, taken from GH, shows, a priori minimax-optimal decision rules are in general different from a posteriori minimax-optimal decision rules.

**Example 2.1:** Suppose that $\mathcal{X} = \mathcal{Y} = \mathcal{A} = \{0, 1\}$ and $\mathcal{P} = \{\Pr \in \Delta(\mathcal{X} \times \mathcal{Y}) : \Pr_\mathcal{Y}(Y = 1) = 2/3\}$. Thus, $\mathcal{P}$ consists of all distributions whose marginal on $Y$ gives $Y = 1$ probability $2/3$. We can think of the actions in $\mathcal{A}$ as predictions of the value of $Y$. The loss function is 0 if the right value is predicted and 1 otherwise; that is, $L(i, j) = |i - j|$. This is the so-called 0/1 or *classification* loss. It is easy to see that the optimal a priori decision rule is to choose 1 no matter what is observed (which has expected loss $1/3$). Intuitively, observing the value of $X$ tells us nothing about the value of $Y$, so the best decision is to predict according to the prior probability of $Y = 1$. However, all probabilities on $Y = 1$ are compatible with observing either $X = 0$ or $X = 1$. That is, both $(\mathcal{P} \mid X = 0)_\mathcal{Y}$ and $(\mathcal{P} \mid X = 1)_\mathcal{Y}$ consist of all distributions on $\mathcal{Y}$. Thus, the minimax optimal a posteriori decision rule randomizes (with equal probability) between $Y = 0$ and $Y = 1$.

Thus, if you make decisions according to the minimax rule, then before making an observation, you will predict $Y = 1$. However, *no matter what observation you make*, after making the observation, you will randomize (with equal probability) between predicting $Y = 0$ and $Y = 1$. Moreover, you know even before making the observation that your opinion as to the best decision rule will change in this way. ∎

## 3 TWO GAME-THEORETIC INTERPRETATIONS OF $\mathcal{P}$

What does it mean that an agent's uncertainty is characterized by a set $\mathcal{P}$ of probability distributions? How should we understand $\mathcal{P}$? We give $\mathcal{P}$ a game-theoretic interpretation here: namely, an adversary gets to choose a distribu-tion from the set $\mathcal{P}$.[3] But this does not completely specify the game. We must also specify *when* the adversary makes the choice. We consider two times that the adversary can choose: the first is before the agents observes the value of $\mathcal{X}$, and the second is after. We formalize this as two different games, where we take the "adversary" to be a bookie.

We call the first game the $\mathcal{P}$-game. It is defined as follows:

1. The bookie chooses a distribution $\Pr \in \mathcal{P}$.
2. The value $x$ of $X$ is chosen (by nature) according to $\Pr_{\mathcal{X}}$ and observed by both bookie and agent.
3. The agent chooses an action $a \in \mathcal{A}$.
4. The value $y$ of $Y$ is chosen according to $\Pr \mid X = x$.
5. The agent's loss is $L(y, a)$; the bookie's loss is $-L(y, a)$.

This is a zero-sum game; the agent's loss is the bookie's gain. In this game, the agent's strategy is a decision rule, that is, a function that gives a distribution over actions for each observed value of $X$. The bookie's strategy is a distribution over distributions in $\mathcal{P}$.

We now consider a second interpretation of $\mathcal{P}$, characterized by a different game that gives the bookie more power. Rather than choosing the distribution before observing the value of $X$, the bookie gets to choose the distribution after observing the value. We call this the $\mathcal{P}$-$X$-game.

1. The value $x$ of $X$ is chosen (by nature) in such a way that $\Pr(X = x) > 0$ for some $\Pr \in \mathcal{P}$, and observed by both the bookie and the agent.
2. The bookie chooses a distribution $\Pr \in \mathcal{P}$ such that $\Pr(X = x) > 0$.[4]
3. The agent chooses an action $a \in \mathcal{A}$.
4. The value $y$ of $Y$ is chosen according to $\Pr \mid X = x$.
5. The agent's loss is $L(y, a)$; the bookie's loss is $-L(y, a)$.

Recall that a pair of strategies $(S_1, S_2)$ is a Nash equilibrium if neither party can do better by unilaterally changing strategies. If, as in our case, $(S_1, S_2)$ is a Nash equilibrium in a zero-sum game, it is also known as a "saddle point"; $S_1$ must be a minimax strategy, and $S_2$ must be a maximin strategy [Grünwald and Dawid 2004]. As the following results show, an agent must be using an a priori minimax-optimal decision rule in a Nash equilibrium of the $\mathcal{P}$-game, and an a posteriori minimax-optimal decision rule is a Nash equilibrium of the $\mathcal{P}$-$X$-game. This can be viewed as a justification for using (a priori and a posteriori) minimax-optimal decision rules.

---
[3] This interpretation remains meaningful in several practical situations where there is no explicit adversary; see the final paragraph of this section.

[4] If we were to consider conditional probability measures, for which $\Pr(Y = y \mid X = x)$ is defined even if $\Pr(X = x) = 0$, then we could drop the restriction that $x$ is chosen such that $\Pr(X = x) > 0$ for some $\Pr \in \mathcal{P}$.

**Theorem 3.1:** *Fix $\mathcal{X}$, $\mathcal{Y}$, $\mathcal{A}$, $L$, and $\mathcal{P} \subseteq \Delta(\mathcal{X} \times \mathcal{Y})$.*

*(a) The $\mathcal{P}$-game has a Nash equilibrium $(\pi^*, \delta^*)$, where $\pi^*$ is a distribution over $\mathcal{P}$ with finite support.*

*(b) If $(\pi^*, \delta^*)$ is a Nash equilibrium of the $\mathcal{P}$-game such that $\pi^*$ has finite support, then*

*(i) for every distribution $\Pr' \in \mathcal{P}$ in the support of $\pi^*$, we have $E_{\Pr'}[L_{\delta^*}] = \max_{\Pr \in \mathcal{P}} E_{\Pr}[L_{\delta^*}]$;*

*(ii) if $\Pr^* = \sum_{\Pr \in \mathcal{P}, \pi^*(\Pr) > 0} \pi^*(\Pr) \Pr$ (i.e., $\Pr^*$ is the convex combination of the distributions in the support of $\pi^*$, weighted by their probability according to $\pi^*$), then*

$$\begin{aligned} E_{\Pr^*}[L_{\delta^*}] &= \min_{\delta \in \mathcal{D}(\mathcal{X}, \mathcal{A})} E_{\Pr^*}[L_\delta] \\ &= \max_{\Pr \in \mathcal{P}} \min_{\delta \in \mathcal{D}(\mathcal{X}, \mathcal{A})} E_{\Pr}[L_\delta] \\ &= \min_{\delta \in \mathcal{D}(\mathcal{X}, \mathcal{A})} \max_{\Pr \in \mathcal{P}} E_{\Pr}[L_\delta] \\ &= \max_{\Pr \in \mathcal{P}} E_{\Pr}[L_{\delta^*}]. \end{aligned}$$

Once nature has chosen a value for $X$ in the $\mathcal{P}$-$X$-game, we can regard steps 2–5 of the $\mathcal{P}$-$X$-game as a game between the bookie and the agent, where the bookie's strategy is characterized by a distribution in $\mathcal{P} \mid X = x$ and the agent's is characterized by a distribution over actions. We call this the $\mathcal{P}$-$x$-game.

**Theorem 3.2:** *Fix $\mathcal{X}$, $\mathcal{Y}$, $\mathcal{A}$, $L$, $\mathcal{P} \subseteq \Delta(\mathcal{X} \times \mathcal{Y})$.*

*(a) The $\mathcal{P}$-$x$-game has a Nash equilibrium $(\pi^*, \delta^*(x))$, where $\pi^*$ is a distribution over $\mathcal{P} \mid X = x$ with finite support.*

*(b) If $(\pi^*, \delta^*(x))$ is a Nash equilibrium of the $\mathcal{P}$-$x$-game such that $\pi^*$ has finite support, then*

*(i) for all $\Pr'$ in the support of $\pi^*$, we have $E_{\Pr'}[L_{\delta^*}] = \max_{\Pr \in \mathcal{P} \mid X = x} E_{\Pr}[L_{\delta^*}]$;*

*(ii) if $\Pr^* = \sum_{\Pr \in \mathcal{P}, \pi^*(\Pr) > 0} \pi^*(\Pr) \Pr$, then*

$$\begin{aligned} E_{\Pr^*}[L_{\delta^*}] &= \min_{\delta \in \mathcal{D}(\mathcal{X}, \mathcal{A})} E_{\Pr^*}[L_\delta] \\ &= \max_{\Pr \in \mathcal{P} \mid X = x} \min_{\delta \in \mathcal{D}(\mathcal{X}, \mathcal{A})} E_{\Pr}[L_\delta] \\ &= \min_{\delta \in \mathcal{D}(\mathcal{X}, \mathcal{A})} \max_{\Pr \in \mathcal{P} \mid X = x} E_{\Pr}[L_\delta] \\ &= \max_{\Pr \in \mathcal{P} \mid X = x} E_{\Pr}[L_{\delta^*}]. \end{aligned}$$

Since all distributions $\Pr$ in the expression $\min_{\delta \in \mathcal{D}(\mathcal{X}, \mathcal{A})} \max_{\Pr \in \mathcal{P} \mid X = x} E_{\Pr}[L_\delta]$ in part (b)(ii) are in $\mathcal{P} \mid X = x$, as in (1), the minimum is effectively over randomized actions rather than decision rules.

The proof of Theorems 3.1 and 3.2, as well as all other missing proofs, can be found in the full paper [Grünwald and Halpern 2007]. These theorems can be viewed as saying that there is no time inconsistency; rather, we must just be careful about what game is being played. If the $\mathcal{P}$-game is being played, the right strategy is the a priori minimax-optimal strategy, both before and after the value of $X$ is observed; similarly, if the $\mathcal{P}$-$X$-game is being played, the right strategy is the a posteriori minimax-optimal strategy, both before and after the value of $X$ is observed. Indeed, thinking in terms of the games explains the apparent time inconsistency. While it is true that the agent gains more information by observing $X = x$, in the $\mathcal{P}$-$X$ game, so does the bookie. This information may be of more use to the bookie than the agent, so, in this game, the agent can be worse off by being given the opportunity to learn the value of $X$.

Of course, in most practical situations, agents (robots, statisticians,...) are not really confronted with a bookie who tries to make them suffer. Rather, the agents may have no idea at all what distribution holds, except that it is in some set $\mathcal{P}$. Because all they know is $\mathcal{P}$, they decide to prepare themselves for the worst-case and play the minimax strategy. The fact that such a minimax strategy can be interpreted in terms of a Nash equilibrium of a game helps to understand differences between different forms of minimax (such as a priori and a posteriori minimax). From this point of view, it seems strange to have a bookie choose between different distributions in $\mathcal{P}$ according to some distribution $\pi^*$. However, if $\mathcal{P}$ is convex, we can replace the distribution $\pi^*$ on $\mathcal{P}$ by a single distribution in $\mathcal{P}$, which consists of the convex combination of the distributions in the support of $\pi^*$; this is just the distribution $\Pr^*$ of Theorems 3.1 and 3.2. Thus, Theorems 3.1 and 3.2 hold with the bookie restricted to a deterministic strategy.

## 4 CHARACTERIZING A PRIORI MINIMAX DECISION RULES

To get the a posteriori minimax-optimal decision rule we do the obvious thing: if $x$ is observed, we simply condition each probability distribution $\Pr \in \mathcal{P}$ on $X = x$, and choose the action that gives the least expected loss (in the worst case) with respect to $\mathcal{P} \mid X = x$.

We might expect that the a priori minimax-optimal decision rule should do the same thing. That is, it should be the decision rule that says, if $x$ is observed, then we choose the action that again gives the best result (in the worst case) with respect to $\mathcal{P} \mid X = x$. However, as shown in GH, this intuition is incorrect in general. There are times, for example, that the best thing to do is to ignore the observed value of $X$, and just choose the action that gives the least expected loss (in the worst case) with respect to $\mathcal{P}$, no matter what value $X$ has. In this section we first give a sufficient condition for conditioning to be optimal, and then characterize when ignoring the observed value is optimal.

**Definition 4.1:** Let $\langle \mathcal{P} \rangle = \{\Pr \in \Delta(\mathcal{X} \times \mathcal{Y}) : \Pr_\mathcal{X} \in \mathcal{P}_\mathcal{X} \text{ and } (\Pr \mid X = x) \in (\mathcal{P} \mid X = x)$ for all $x \in \mathcal{X}$ such that $\mathcal{P} \mid X = x$ is nonempty$\}$. ∎

Thus, $\langle \mathcal{P} \rangle$ consists of all distributions $\Pr$ whose marginal on $\mathcal{X}$ is the marginal on $\mathcal{X}$ of some distribution in $\mathcal{P}$ and whose conditional on observing $X = x$ is the conditional of some distribution in $\mathcal{P}$, for all $x \in \mathcal{X}$. Clearly $\mathcal{P} \subseteq \langle \mathcal{P} \rangle$, but the converse is not necessarily true. When it is true, conditioning is optimal.

**Proposition 4.2:** *If $\mathcal{P} = \langle \mathcal{P} \rangle$, then there exists an a priori minimax-optimal rule that is also a posteriori minimax optimal. If, for all $\Pr \in \mathcal{P}$ and all $x \in \mathcal{X}$, $\Pr(X = x) > 0$, then every a priori minimax-optimal rule is also a posteriori minimax optimal.*

As we saw in Example 2.1, the minimax-optimal a priori decision rule is not always the same as the minimax-optimal a posteriori decision rule. In fact, the minimax-optimal a priori decision rule ignores the information observed. Formally, a rule $\delta$ *ignores information* if $\delta(x) = \delta(x')$ for all $x, x' \in \mathcal{X}$. If $\delta$ ignores information, define $L'_\delta$ to be the random variable on $\mathcal{Y}$ such that $L'_\delta(y) = L_\delta(x, y)$ for some choice of $x$. This is well defined, since $L_\delta(x, y) = L_\delta(x', y)$ for all $x, x' \in \mathcal{X}$.

**Theorem 4.3:** *Fix $\mathcal{X}$, $\mathcal{Y}$, $L$, $\mathcal{A}$, and $\mathcal{P} \subseteq \Delta(\mathcal{X} \times \mathcal{Y})$. If, for all $\Pr_\mathcal{Y} \in \mathcal{P}_\mathcal{Y}$, $\mathcal{P}$ contains a distribution $\Pr'$ such that $X$ and $Y$ are independent under $\Pr'$, and $\Pr'_\mathcal{Y} = \Pr_\mathcal{Y}$, then there is an a priori minimax-optimal decision rule that ignores information. Under these conditions, if $\delta$ is an a priori minimax-optimal decision rule that ignores information, then $\delta$ essentially optimizes with respect to the marginal on $Y$; that is, $\max_{\Pr \in \mathcal{P}} E_{\Pr}[L_\delta] = \max_{\Pr_\mathcal{Y} \in \mathcal{P}_\mathcal{Y}} E_{\Pr_\mathcal{Y}}[L'_\delta]$.*

GH focused on the case that $\mathcal{P}_\mathcal{Y}$ is a singleton (i.e., the marginal probability on $Y$ is the same for all distributions in $\mathcal{P}$) and for all $x$, $\mathcal{P}_\mathcal{Y} \subseteq (\mathcal{P} \mid X = x)_\mathcal{Y}$. It is immediate from Theorem 4.3 that ignoring information is a priori minimax optimal in this case.

## 5 $\mathcal{C}$-CONDITIONING & CALIBRATION

Conditioning is the most common way of updating uncertainty. In this section, we examine updating by conditioning. The following definition makes precise the idea that a decision rule is based on conditioning.

**Definition 5.1:** A *probability update rule* is a function $\Pi : 2^{\Delta(\mathcal{X} \times \mathcal{Y})} \times \mathcal{X} \to 2^{\Delta(\mathcal{X} \times \mathcal{Y})}$ mapping a set $\mathcal{P}$ of distributions and an observation $x$ to a set $\Pi(\mathcal{P}, x)$ of distributions; intuitively, $\Pi(\mathcal{P}, x)$ is the result of updating $\mathcal{P}$ with the observation $x$. ∎

**Definition 5.2:** Let $\mathcal{C} = \{\mathcal{X}_1, \ldots, \mathcal{X}_k\}$ be a partition of $\mathcal{X}$; that is, $\mathcal{X}_i \neq \emptyset$ for $i = 1, \ldots, k$; $\mathcal{X}_1 \cup \ldots \mathcal{X}_k = \mathcal{X}$; and $\mathcal{X}_i \cap \mathcal{X}_j = \emptyset$ for $i \neq j$. If $x \in \mathcal{X}$, let $\mathcal{C}(x)$ be the cell containing $x$; i.e., the unique element $\mathcal{X}_i \in \mathcal{C}$ such that $x \in \mathcal{X}_i$. The $\mathcal{C}$-*conditioning* probability update rule is the function $\Pi$ defined by taking $\Pi(\mathcal{P}, x) = \mathcal{P} \mid X \in$

$\mathcal{C}(x)$. A decision rule $\delta$ is *based on $\mathcal{C}$-conditioning* if it amounts to first updating the set $\mathcal{P}$ to $\mathcal{P} \mid X \in \mathcal{C}(x)$, and then taking the minimax-optimal distribution over actions relative to $\mathcal{P} \mid X \in \mathcal{C}(x)$. Formally, $\delta$ is based on $\mathcal{C}$-conditioning if, for all $x \in \mathcal{X}$ with $\Pr(X = x) > 0$ for some $\Pr \in \mathcal{P}$,

$$\max_{\Pr \in \mathcal{P} \mid X \in \mathcal{C}(x)} E_{\Pr}[L_\delta] = \min_{\delta \in \mathcal{D}(\mathcal{X}, \mathcal{A})} \max_{\Pr \in \mathcal{P} \mid X \in \mathcal{C}(x)} E_{\Pr}[L_\delta].$$

∎

All examples of a priori minimax decision rules that we have seen so far are based on $\mathcal{C}$-conditioning: Standard conditioning is based on $\mathcal{C}$-conditioning, where we take $\mathcal{C}$ to consist of all singletons; ignoring information is also based on $\mathcal{C}$-conditioning, where $\mathcal{C} = \{\mathcal{X}\}$. This suggests that, perhaps, the a priori minimax decision rule must also be based on $\mathcal{C}$-conditioning. The following well-known example shows that this conjecture is false.

**Example 5.3:** [**The Monty Hall Problem**] [Mosteller 1965; vos Savant 1990]: Suppose that you're on a game show and given a choice of three doors. Behind one is a car; behind the others are goats. You pick door 1. Before opening door 1, Monty Hall, the host (who knows what is behind each door) opens one of the other two doors, say, door 3, which has a goat. He then asks you if you still want to take what's behind door 1, or to take what's behind door 2 instead. Should you switch? You may assume that initially, the car was equally likely to be behind each of the doors.

We formalize this well-known problem as a $\mathcal{P}$-game, as follows: $\mathcal{Y} = \{1, 2, 3\}$ represents the door which the car is behind. $\mathcal{X} = \{G_2, G_3\}$, where, for $j \in \{2, 3\}$, $G_j$ corresponds to the quizmaster showing that there is a goat behind door $j$. $\mathcal{A} = \{1, 2, 3\}$, where action $a \in \mathcal{A}$ corresponds to the door you finally choose, after Monty has opened door 2 or 3. The loss function is once again the classification loss, $L(i, j) = 1$ if $i \neq j$, that is, if you choose a door with a goat behind it, and $L(i, j) = 0$ if $i = j$, that is, if you choose a door with a car. $\mathcal{P}$ is the set of all distributions $\Pr$ on $\mathcal{X} \times \mathcal{Y}$ satisfying

$$\Pr_{\mathcal{Y}}(Y = 1) = \Pr_{\mathcal{Y}}(Y = 2) = \Pr_{\mathcal{Y}}(Y = 3) = \tfrac{1}{3}$$
$$\Pr(Y = 2 \mid X = G_2) = \Pr(Y = 3 \mid X = G_3) = 0.$$

It is well known, and easy to show, that the minimax-optimal strategy is always to switch doors, no matter whether Monty opens door 2 or door 3. Since the game is an instance of the $\mathcal{P}$-game, this means that the decision rule $\delta^*$ given by $\delta^*(G_2) = 3$ ; $\delta^*(G_3) = 2$ is an a priori minimax rule. Clearly, $\delta^*$ is *not* based on $\mathcal{C}$-conditioning: there exist only two partitions of $\mathcal{X}$. The corresponding two update rules based on $\mathcal{C}$-conditioning amount to, respectively, (a) ignoring $X$ and choosing each door with probability 1/3, or (b) conditioning on $X$ in the standard way and thus choosing each of the two remaining doors with probability 1/2. Neither strategy (a) nor (b) is minimax optimal. Thus, the a priori minimax decision rule in the $\mathcal{P}$-game is not always based on $\mathcal{C}$-conditioning. ∎

While the example shows that $\mathcal{C}$-conditioning is not always optimal in the minimax sense, it can be justified by other means; as we now show, $\mathcal{C}$-conditioning is closely related to *calibration*. Indeed, a probability update rule is calibrated if and only if for each $\mathcal{P}$, it amounts to $\mathcal{C}$-conditioning for some partition $\mathcal{C}$ of $\mathcal{X}$. Calibration is usually defined in terms of empirical data. To explain what it means, consider a weather forecaster, who predicts the probability of rain every day. How should we interpret the probabilities that she announces? The usual interpretation—which coincides with most people's intuitive understanding—is that, in the long run, on those days at which the weather forecaster predict probability $p$, it will rain approximately $100p\%$ of the time [Dawid 1982]. Thus, for example, among all days for which she predicted 0.1, the fraction of days with rain was close to 0.1 (given the weather forecaster's precision, we should require it to be between, say, 0.05 and 0.15). A weather forecaster with this property is said to be *calibrated*. If a weather forecaster is calibrated, and you make bets based on her probabilistic predictions (which are all accepted), then in the long run you will not lose money.

If a weather forecaster is not calibrated, there exist bets which seem favorable but which result in a loss. Note that calibration is a *minimal* requirement: a weather forecaster who predicts 0.3 for every single day of the year may be calibrated if it indeed rains on 30% of the days, but still not very informative. Thus, given two calibrated forecasters, we prefer the one that makes "sharper" predictions, in a sense to be defined below.

In our case, we do not test probabilistic predictions with respect to empirical relative frequencies, but with respect to other sets of "potentially underlying" probability measures. We are not the first to do this; see, for example, [Vovk, Gammerman, and Shafer 2005]. The definition of calibration extends naturally to this situation. To see how, we first define calibration with respect to a single underlying probability measure. Let $\mathcal{P} = \{\Pr\}$ for a single distribution $\Pr$ and let $\Pi$ be a probability update rule (Definition 5.1) such that $\Pi(\{\Pr\}, x)$ contains just a single distribution for each $x \in \mathcal{X}$ (for example, $\Pi$ could be ordinary conditioning). We define

$$\mathbf{R} = \{\mathcal{R} : \mathcal{R} = (\,\Pi(\mathcal{P}, x)\,)_{\mathcal{Y}} \text{ for some } x \in \mathcal{X}\}. \quad (2)$$

$\mathbf{R}$ is just the range of $\Pi$, restricted to distributions of $Y$, the random variable that we are interested in predicting; its elements are the distributions on $Y$ that $\Pr$ is mapped to, upon observing different values of $x$. Note that $\mathbf{R}$ is defined relative to a probability update rule $\Pi$ and a set $\mathcal{P}$ of distributions. By our assumptions on $\mathcal{P}$ and $\Pi$,

$\mathcal{R} = \{\{R_1\}, \{R_2\}, \ldots\}$ is a set of singleton sets, each containing one distribution on $\mathcal{Y}$. For $\{R\} \in \mathbf{R}$, let $\mathcal{X}_R$ be the set of $x \in \mathcal{X}$ that map Pr to $R$, i.e.

$$\mathcal{X}_R = \{x \in \mathcal{X} \ : \ (\ \Pi(\{\Pr\}, x)\ )_\mathcal{Y} = \{R\}\}.$$

Note that the sets $\{\mathcal{X}_R\}$ partition $\mathcal{X}$. $\Pi$ is calibrated relative to $\mathcal{P}$ if for all $R$ with $\{R\} \in \mathbf{R}$, $(\Pr \mid X \in \mathcal{X}_R)_\mathcal{Y} = R$. Thus, conditioned on the event that the agent predicts $Y$ using distribution $R$, the distribution of $Y$ must indeed be equal to $R$.

It is straightforward to generalize this notion to sets $\mathcal{P}$ of probability distributions that are not singletons, and update rules $\Pi$ that map to sets of probabilities. Definition (2) remains unchanged. For $\mathcal{R} \in \mathbf{R}$, we now take $\mathcal{X}_\mathcal{R}$ to be the set of $x \in \mathcal{X}$ that map $\mathcal{P}$ to $\mathcal{R}$, that is,

$$\mathcal{X}_\mathcal{R} = \{x \in \mathcal{X} \ : \ (\ \Pi(\mathcal{P}, x)\ )_\mathcal{Y} = \mathcal{R}\}. \quad (3)$$

Once again, the sets $\{\mathcal{X}_\mathcal{R}\}$ partition $\mathcal{X}$.

**Definition 5.4:** $\Pi$ is *calibrated relative to* $\mathcal{P}$ if for all $\Pr \in \mathcal{P}$ and $\mathcal{R} \in \mathbf{R}$, $\Pr_Y(\cdot \mid X \in \mathcal{X}_\mathcal{R}) \in \mathcal{R}$.
$\Pi$ *is calibrated* if it is calibrated relative to all sets of distributions $\mathcal{P} \subseteq \Delta(\mathcal{X} \times \mathcal{Y})$. ∎

**Proposition 5.5:** *For all partitions $\mathcal{C}$ of $\mathcal{X}$ and all $\mathcal{P}$, $\mathcal{C}$-conditioning is calibrated relative to $\mathcal{P}$.*

Calibration as defined here is a very weak notion. For example, the update rule $\Pi(\mathcal{P}, x) = \Delta(\mathcal{X} \times \mathcal{Y})$ that maps each combination of $x$ and $\mathcal{P}$ to the set of all distributions on $\mathcal{X} \times \mathcal{Y}$ is calibrated under our definition. This update rule loses whatever information may have been contained in $\mathcal{P}$, and is therefore not very useful. Intuitively, the fewer distributions that there are in $\mathcal{P}$, the more information $\mathcal{P}$ contains. Thus, we restrict ourselves to sets $\mathcal{P}$ that are as small as possible, while still being calibrated.

**Definition 5.6:** Update rule $\Pi'$ is *wider than update rule* $\Pi$ *relative to* $\mathcal{P}$ if, for all $x \in \mathcal{X}$, $\Pi(\mathcal{P}, x) \subseteq \Pi'(\mathcal{P}, x)$.
$\Pi'$ is *strictly wider* relative to $\mathcal{P}$ if the inclusion is strict for some some $x$. $\Pi$ is *(strictly) narrower* than $\Pi'$, relative to $\mathcal{P}$ if $\Pi'$ is (strictly) wider than $\Pi$ relative to $\mathcal{P}$. $\Pi$ is *sharply calibrated* relative to $\mathcal{P}$ if $\Pi$ is calibrated relative to $\mathcal{P}$ and there is no update rule $\Pi'$ that is calibrated and strictly narrower than $\Pi$ relative to $\mathcal{P}$. $\Pi$ is *sharply calibrated* if $\Pi$ is sharply calibrated relative to all $\mathcal{P} \subseteq \Delta(\mathcal{X} \times \mathcal{Y})$. ∎

We now want to prove that every sharply calibrated update rule must involve conditioning. To make this precise, we need the following definition.

**Definition 5.7:** $\Pi$ is a *generalized conditioning update rule* if, for all $\mathcal{P} \subseteq \Delta(\mathcal{X} \times \mathcal{Y})$, there exists a partition $\mathcal{C}$ (that may depend on $\mathcal{P}$) such that for all $x \in \mathcal{X}$, $\Pi(\mathcal{P}, x) = \mathcal{P} \mid \mathcal{C}(x)$. ∎

Note that in a generalized conditioning rule, we condition on a partition of $\mathcal{X}$, but the partition may depend on the set $\mathcal{P}$. For example, for some $\mathcal{P}$, the rule may ignore the value of $x$, whereas for other $\mathcal{P}$, it may amount to ordinary conditioning. It easily follows from Proposition 4.2 that every generalized conditioning rule is calibrated. The next result shows that every *sharply* calibrated update rule must be a generalized conditioning rule.

**Theorem 5.8:** *There exists an update rule that is sharply calibrated. Moreover, every sharply calibrated update rule is a generalized conditioning update rule.*

Theorem 5.8 says that an agent who wants to be sharply calibrated should update her probabilities using conditioning (although what she conditions on may depend on the set of probabilities that she considers possible).

Given the game-theoretic interpretation of Section 3, we might wonder if there is a variant of the games considered earlier for which the equilibrium involves generalized conditioning. As we show in the full paper, there is (although the game is perhaps not as natural as the ones considered in Section 3). Roughly speaking, we consider a three-player game, with a bookie and two agents. The bookie again chooses a probability distribution from a set $\mathcal{P}$; the bookie also chooses the loss function from some set. The first agent observes $\mathcal{P}$ and $x$ and updates $\mathcal{P}$ to $\mathcal{P}_x$. The second agent learns $\mathcal{P}_x$ and $b$ (but not $\mathcal{P}$ and $x$) and makes the minimax-optimal decision. As we show, in Nash equilibrium, the first agent's updated set of probabilities, $\mathcal{P}_x$, must be the result of $\mathcal{C}$-conditioning, where, as in Theorem 5.8, $\mathcal{C}$ may depend on $\mathcal{P}$.

## 6 DISCUSSION

We have examined how to update uncertainty represented by a set of probability distributions, where we motivate updating rules in terms of the minimax criterion. Our key innovation has been to show how different approaches can be understood in terms of a game between a bookie and an agent, where the bookie picks a distribution from the set and the agent chooses an action after making an observation. Different approaches to updating arise depending on whether the bookie's choice is made before or after the observation. We believe that this game-theoretic approach should prove useful more generally in understanding different approaches to updating. We hope to explore this further in future work.

We end this paper by giving an overview of the senses in which conditioning is optimal and the senses in which it is not, when uncertainty is represented by a set of distributions. We have established that conditioning the full set $\mathcal{P}$ on $X = x$ is minimax optimal in the $\mathcal{P}$-$x$-game, but not in the $\mathcal{P}$-game. The minimax-optimal decision rule in the

$\mathcal{P}$-game is often an instance of $\mathcal{C}$-conditioning, a generalization of conditioning. The Monty Hall problem showed, however, that this is not always the case. On the other hand, if instead of the minimax criterion, we insist that update rules are calibrated, then $\mathcal{C}$-conditioning is always the right thing to do after all.

There are two more senses in which conditioning is the right thing to do. First, Walley [1991] shows that, in a sense, conditioning is the only updating rule that is *coherent*, according to his notion of coherence. He justifies coherence decision theoretically, but not by using the minimax criterion. Note that the minimax criterion puts a total order on decision rules. That is, we can say that $\delta$ is at least as good as $\delta'$ if

$$\max_{\Pr \in \mathcal{P}} E_{\Pr}[L_\delta] \leq \max_{\Pr \in \mathcal{P}} E_{\Pr}[L_{\delta'}].$$

By way of contrast, Walley [1991] puts a partial order on decision rules by taking $\delta$ to be at least as good as $\delta'$ if

$$\max_{\Pr \in \mathcal{P}} E_{\Pr}[L_\delta - L_{\delta'}] \leq 0.$$

Since both $\max_{\Pr \in \mathcal{P}} E_{\Pr}[L_\delta - L_{\delta'}]$ and $\max_{\Pr \in \mathcal{P}} E_{\Pr}[L_{\delta'} - L_\delta]$ may be positive, this is indeed a partial order. If we use this ordering to determine the optimal decision rule then, as Walley shows, conditioning is the only right thing to do.

Second, in this paper, we interpreted "conditioning" as conditioning the full given set of distributions $\mathcal{P}$. Then conditioning is not always an a priori minimax optimal strategy on the observation $X = x$. Alternatively, we could first somehow select a *single* $\Pr \in \mathcal{P}$, condition $\Pr$ on the observed $X = x$, and then take the optimal action relative to $\Pr \mid X = x$. It follows from Theorem 3.1 that the minimax-optimal decision rule $\delta^*$ in a $\mathcal{P}$-game can be understood this way. It defines the optimal response to the distribution $\Pr^* \in \Delta(\mathcal{X} \times \mathcal{Y})$ defined in Theorem 3.1(b)(ii). If $\mathcal{P}$ is convex, then $\Pr^* \in \mathcal{P}$. In this sense, the minimax-optimal decision rule can always be viewed as an instance of "conditioning," but on a single special $\Pr^*$ that depends on the loss function $L$ rather than on the full set $\mathcal{P}$.

It is worth noting that Grove and Halpern [1998] give an axiomatic characterization of conditioning sets of probabilities, based on axioms given by van Fraassen [1987, 1985] that characterizing conditioning in the case that uncertainty is characterized by a single probability measure. As Grove and Halpern point out, their axioms are not as compelling as those of van Fraassen. It would be interesting to know whether a similar axiomatization can be used to characterize the update notions that we have considered here.

**Acknowledgments** Peter Grünwald was supported by the IST Programme of the European Community, under the PASCAL Network of Excellence, IST-2002-506778. Joseph Halpern was supported in part by NSF under grants ITR-0325453 and IIS-0534064, and by AFOSR under grant FA9550-05-1-0055.

## References

Augustin, T. (2003). On the suboptimality of the generalized Bayes rule and robust Bayesian procedures from the decision theoretic point of view. In *Proceedings ISIPTA '03*, pp. 31–45.

Cozman, F. G. and P. Walley (2001). Graphoid properties of epistemic irrelevance and independence. In *Proceedings ISIPTA '01*, pp. 112–121.

Dawid, A. (1982). The well-calibrated Bayesian. *Journal of the American Statistical Association 77*.

Gärdenfors, P. and N. Sahlin (1982). Unreliable probabilities, risk taking, and decision making. *Synthese 53*, 361–386.

Gilboa, I. and D. Schmeidler (1989). Maxmin expected utility with a non-unique prior. *Journal of Mathematical Economics 18*, 141–153.

Grove, A. J. and J. Y. Halpern (1998). Updating sets of probabilities. In *Proceedings UAI '98*, pp. 173–182.

Grünwald, P. and A. Dawid (2004). Game theory, maximum entropy, minimum discrepancy, and robust Bayesian decision theory. *The Annals of Statistics 32*(4), 1367–1433.

Grünwald, P. and J. Halpern (2004). When ignorance is bliss. In *Proceedings UAI '04*, pp. 226–1.734.

Grünwald, P. and J. Halpern (2007). A game-theoretic analysis of updating sets of probabilities. http://arxiv.org/abs/0711.3235.

Herron, T., T. Seidenfeld, and L. Wasserman (1997). Divisive conditioning: Further results on dilation. *Philosophy of Science 64*, 411–444.

Hughes, R. I. G. and B. C. van Fraassen (1985). Symmetry arguments in probability kinematics. In P. Kitcher and P. Asquith (Eds.), *PSA 1984*, Volume 2, pp. 851–869. Philosophy of Science Association.

Mosteller, F. (1965). *Fifty Challenging Problems in Probability with Solutions*. Addison-Wesley.

Seidenfeld, T. (2004). A contrast between two decision rules for use with (convex) sets of probabilities: $\gamma$-maximin versus $E$-admissibility. *Synthese*.

Seidenfeld, T. and L. Wasserman (1993). Dilation for convex sets of probabilities. *Annals of Statistics 21*, 1139–1154.

van Fraassen, B. C. (1987). Symmetries of personal probability kinematics. In N. Rescher (Ed.), *Scientific Enquiry in Philsophical Perspective*, pp. 183–223. University Press of America.

vos Savant, M. (Sept. 9, 1990). Ask Marilyn. *Parade Magazine*, 15.

Vovk, V., A. Gammerman, and G. Shafer (2005). *Algorithmic Learning in a Random World*. Springer.

Wald, A. (1950). *Statistical Decision Functions*. Wiley.

Walley, P. (1991). *Statistical Reasoning with Imprecise Probabilities*. Chapman and Hall.